\documentclass[runningheads]{llncs}

\usepackage{amsmath}
\usepackage{amssymb}
\usepackage[misc]{ifsym}
\usepackage{graphicx}
\usepackage{color}

\usepackage{algorithm}
\usepackage{algorithmicx}
\usepackage{algpseudocode}
\usepackage{multirow}
\usepackage{array}
\newcolumntype{L}[1]{>{\raggedright\let\newline\\\arraybackslash\hspace{0pt}}m{#1}}
\newcolumntype{C}[1]{>{\centering\let\newline\\\arraybackslash\hspace{0pt}}m{#1}}
\newcolumntype{R}[1]{>{\raggedleft\let\newline\\\arraybackslash\hspace{0pt}}m{#1}}

\usepackage{url}
\urldef{\mailsa}\path|{alfred.hofmann, ursula.barth, ingrid.haas, frank.holzwarth,|
\urldef{\mailsb}\path|anna.kramer, leonie.kunz, christine.reiss, nicole.sator,|
\urldef{\mailsc}\path|erika.siebert-cole, peter.strasser, lncs}@springer.com|
\newcommand{\keywords}[1]{\par\addvspace\baselineskip
\noindent\keywordname\enspace\ignorespaces#1}

\usepackage{longtable}
\usepackage{booktabs}
\usepackage{bm}
\begin{document}

\mainmatter  % start of an individual contribution

% first the title is needed
\title{Multi-Scale Coarse-to-Fine Segmentation for Screening Pancreatic Ductal Adenocarcinoma}

% a short form should be given in case it is too long for the running head
\titlerunning{Multi-Scale Coarse-to-Fine Segmentation for Screening PDAC}

% the name(s) of the author(s) follow(s) next
%
% NB: Chinese authors should write their first names(s) in front of
% their surnames. This ensures that the names appear correctly in
% the running heads and the author index.
%
%%%% BEGIN %%%%
% \author{Anonymous Authors}
%
\authorrunning{Z. Zhu \emph{et al.}}
% (feature abused for this document to repeat the title also on left hand pages)

% the affiliations are given next; don't give your e-mail address
% unless you accept that it will be published
% \institute{Anonymous Institute}
%%%% END %%%%

\author{Zhuotun Zhu\textsuperscript{*,1}, Yingda Xia\textsuperscript{*,1}, \\
Lingxi Xie\textsuperscript{1}, Elliot K. Fishman\textsuperscript{2}, Alan L. Yuille\textsuperscript{1}}
%index{Zhu, Zhuotun}
%index{Yingda, Xia}
%index{Lingxi, Xie}
%index{Fishman, Elliot}
%index{Yuille, Alan}

\institute{
\textsuperscript{1}The Johns Hopkins University, Baltimore, MD 21218, USA\\
{\tt\small \{zhuotun,philyingdaxia,198808xc,alan.l.yuille\}@gmail.com}\\
\textsuperscript{2}The Johns Hopkins University School of Medicine, Baltimore, MD 21287, USA\\
% {\tt\small zhuotun@gmail.com}\quad
% {\tt\small philyingdaxia@gmail.com}\quad
% {\tt\small 198808xc@gmail.com}\quad
% {\tt\small efishman@jhmi.edu}\quad
% {\tt\small alan.l.yuille@gmail.com}\\
{\tt\small efishman@jhmi.edu}\\
}

%
% NB: a more complex sample for affiliations and the mapping to the
% corresponding authors can be found in the file "llncs.dem"
% (search for the string "\mainmatter" where a contribution starts).
% "llncs.dem" accompanies the document class "llncs.cls".
%

%\toctitle{Multi-Scale Coarse-to-Fine Segmentation for Screening PDAC}
%\tocauthor{Anonymous Authors}
\maketitle

\let\thefootnote\relax\footnote{\text{*} The first two authors equally contributed to the work.}

\begin{abstract}
We propose an intuitive approach of detecting pancreatic ductal adenocarcinoma (PDAC), the most common type of pancreatic cancer, by checking abdominal CT scans. Our idea is named {\bf multi-scale segmentation-for-classification}, which classifies volumes by checking if at least a sufficient number of voxels is segmented as tumors, by which we can provide radiologists with tumor locations. In order to deal with tumors with different scales, we train and test our volumetric segmentation networks with multi-scale inputs in a coarse-to-fine flowchart. A post-processing module is used to filter out outliers and reduce false alarms. We collect a new dataset containing 439 CT scans, in which $136$ cases were diagnosed with PDAC and $303$ cases are normal, which is the largest set for PDAC tumors to the best of our knowledge. To offer the best trade-off between sensitivity and specificity, our proposed framework reports a sensitivity of $94.1\%$ at a specificity of $98.5\%$, which demonstrates the potential to make a clinical impact.

% Pancreatic Ductal Adenocarcinoma
\end{abstract}

% \vspace{-1.0cm}
\section{Introduction}
% \vspace{-0.2cm}
\label{Introduction}

Pancreatic cancer is one of the most dangerous killers to human lives, causing more than $330\rm{,}000$ deaths globally in 2014~\cite{stewart2017world}. Pancreatic ductal adenocarcinoma (PDAC) is the most common type of pancreatic cancer, accounting for about $85\%$ of cancer cases. In early stages, this disease often has few symptoms and is very difficult to discover. By the time of diagnosis, the cancer has often spread to other parts of the body, leading to a very poor prognosis ({\em e.g.}, a five-year survival rate of $5\%$~\cite{stewart2017world}). But, for cases diagnosed early, the survival rate rises to about $20\%$~\cite{board2017pancreatic}. Hence, it is very important to study the possibility of detecting PDAC in common examinations, {\em e.g.}, the abdominal CT scan.

The early diagnosis of pancreatic cancer requires much expertise in reading the scanned images and making decisions, but the increasing number of cases makes it impossible for a limited number of experienced radiologists to check all CT scans manually. Therefore, an artificial intelligence system for this purpose is in need. In particular, the radiologists in our team are interested in a system working on abdominal CT scans, which filters out a large fraction of normal cases, but preserves almost all abnormal cases for further investigation. To the best of our knowledge, there is no existing work on this task.

% \cite{he2016deep}
% This problem falls into the area of computer-assisted diagnosis (CAD). 
% in particular the state-of-the-art convolutional neural networks for image recognition

With the development of deep learning~\cite{krizhevsky2012imagenet}, it is possible to construct a system which learns from professional knowledge in data annotation, and apply it to helping doctors in various clinical purposes. The pancreas is one of the most challenging organs in CT segmentation~\cite{roth2015deeporgan}. The difficulty mainly lies in its irregular shape and low contrast around the boundary. Powered by the recent progress in deep learning for 2D~\cite{chen2016deeplab}\cite{ronneberger2015u} and 3D~\cite{milletari2016v}\cite{xia2018bridging} image segmentation, researchers designed various approaches~\cite{roth2016spatial}\cite{zhu2018a} towards accurate pancreas segmentation. In the pathological cases, the morphology of the pancreas can be largely impacted by the difference in the pancreatic cancer stage~\cite{zhang2017personalized}\cite{zhou2017deep}.

% , making it more difficult to segment the pancreas and lesion areas accurately~\cite{zhou2017fixed}.

% \cite{zhu2016deep}

\begin{figure}[!t]
\begin{center}
    \includegraphics[width=0.9\linewidth]{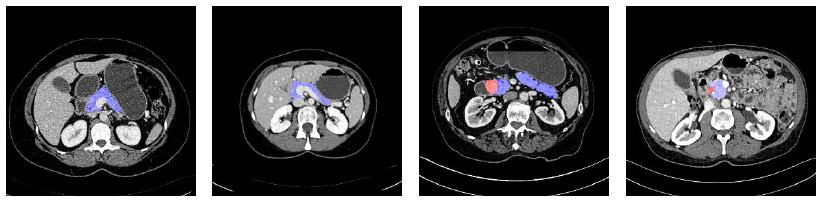}
\end{center}
% \vspace{-0.8cm}
\caption{
    Examples of normal and abnormal (PDAC) pancreases (best viewed in color). Blue and red region mark the normal pancreas and tumor regions, respectively.
    % The tumor sizes vary a lot in the PDAC cases.
    %  (we show a medium-sized one and a small-sized one)
}
\label{Fig:Problem}
% \vspace{-0.8cm}
\end{figure}

% ~\cite{dou2017multilevel}\cite{hussein2018supervised}
Our work is aimed to detect PDAC from a mixture of normal and abnormal CT scans. This is not a simple classification since radiologists also want to know the location of PDAC, we suggest a solution named {\bf segmentation-for-classification}, which trains segmentation models and uses their outputs for classification. To deal with tumors of various sizes (Fig.~\ref{Fig:Problem}), we adopt a segmentation network with multiple input scales, {\em i.e.}, $64^3$, $32^3$ and $16^3$ volumes. But, voting that small input regions lead to a high false alarm rate, we adopt a {\bf coarse-to-fine} testing strategy, which uses the $64^3$ network for a coarse scan, and then the $32^3\&16^3$ networks inside the bounding box to detect small tumors that are possibly ignored in the previous stage. A non-parameterized post-processing algorithm is designed to remove outliers. % A testing volume is classified as PDAC if at least $50$ voxels are segmented as tumor.

% \textcolor{red}{Rewrite this paragraph explicitly with our contributions.}
% We perform experiments on our own dataset with a mixture of normal and abnormal CT scans. In tumor segmentation, our multi-scale approach achieves an average DSC of $56.46\%$ over $136$ cases. In classification, we miss $8$ out of $136$ abnormal cases (sensitivity is $94.1\%$), and false-alarm $3$ of $200$ normal cases (specificity is $98.5\%$). Both the classification and segmentation results can assist the radiologists in further investigation, and largely reduce their workload.

Our contributions are three folds: $1$) we voxelwisely annotate an abdominal CT dataset with $439$ cases in total, in which $136$ cases are diagnosed with PDAC while the remaining $303$ cases are normal, which is currently the largest PDAC dataset to the best of our knowledge; $2$) we adopt a {\bf multi-scale segmentation-for-classification} framework to conduct an {\bf interpretable} abnormality detection, which provides radiologists with suspicious regions for further diagnosis; $3$) our framework achieves a sensitivity of $94.1\%$ at a specificity of $98.5\%$,  which shows a promising direction to make a potential significant clinical impact.

% \vspace{-0.3cm}
\section{The Segmentation-for-Classification Approach}
\label{Approach}

% \vspace{-0.2cm}
\subsection{The Overall Framework}
\label{Approach:Framework}

Let a dataset be ${\mathbf{S}}={\left\{\left(\mathbf{X}_1,y_1^\star\right),\ldots,\left(\mathbf{X}_N,y_N^\star\right)\right\}}$, where $N$ is the number of CT scans, ${\mathbf{X}_n}\in{\mathbb{R}^{W_n\times H_n\times L_n}}$ is the 3D volume with each element indicating the Hounsfield unit (HU) of a voxel, and ${y_n}\in{\left\{0,1\right\}}$ is the label ($0$ for a normal case, $1$ for an abnormal case). Throughout this paper, by {\em abnormal} we refer to the cases diagnosed as PDAC. The goal is to design a model $\mathbb{M}:{y}={f\!\left(\mathbf{X}\right)}$ to predict the label for each testing volume. We evaluate our approach by ranking all volumes by the probability of being a PDAC, computing the sensitivity and specificity at a given threshold, and plotting the ROC curve indicating the relationship between the sensitivity and specificity at different thresholds. For clinical purposes, we shall guarantee a high sensitivity with a reasonable specificity.

%  (few PDAC cases are missed) (there are not too many false alarms).

% (our approach can be combined with other cues such as the shape of the pancreas)
Although some previous work suggested to classify CT or MRI volumes directly using 3D networks~\cite{dou2017multilevel}\cite{hussein2018supervised}, we argue that a better solution is to perform tumor segmentation at the same time of classification. This makes the classification results {\bf interpretable} by segmentation cues, by which radiologists can take a further investigation of the suspicious abnormal regions. In addition, this integrates voxel-wise annotations into the classification model as deep supervision, so that the entire network is better trained~\cite{zhou2017deep}. Therefore, we propose a two-stage framework named {\em segmentation-for-classification}, in which a segmentation stage first extracts voxel-wise cues from the input CT scan, and a classification stage follows to summarize this information into the final prediction. Our multi-scale segmentation strategy is different from~\cite{zhu2018a}, which applied another network of the same scale in the fine stage. {\bf Tumor detection requires multiple scales.}

Mathematically, let each training data be augmented by a segmentation mask $\mathbf{M}_n^\star$ of the same dimensionality as $\mathbf{X}$, so that ${m_{n,i}^\star}\in{\left\{0,1,2\right\}}$ indicates the category of the $i$-th voxel, {\em i.e.}, in the tumor (${m_{n,i}}={2}$), outside the tumor but inside the pancreas (${m_{n,i}}={1}$), or outside the pancreas (${m_{n,i}}={0}$). Note that the tumor voxel set is a subset of the pancreas voxel set. The segmentation module is a high-dimensional function ${\mathbf{M}}={\mathbf{s}\!\left(\mathbf{X}\right)}$, which is implemented by a deep encoder-decoder network. The classification module is a binary function ${y}={c\!\left(\mathbf{M}\right)}$. The overall framework is thus written as:
\begin{equation}
\label{Eqn:Framework}
{y}={f\!\left(\mathbf{X}\right)}={c\circ\mathbf{s}\!\left(\mathbf{X}\right)}.
\end{equation}

% \vspace{-0.5cm}
\subsection{Training: Multi-Scale Deeply-Supervised Segmentation}
\label{Approach:Training}

\begin{figure}[!t]
\begin{center}
    \includegraphics[width=12.0cm]{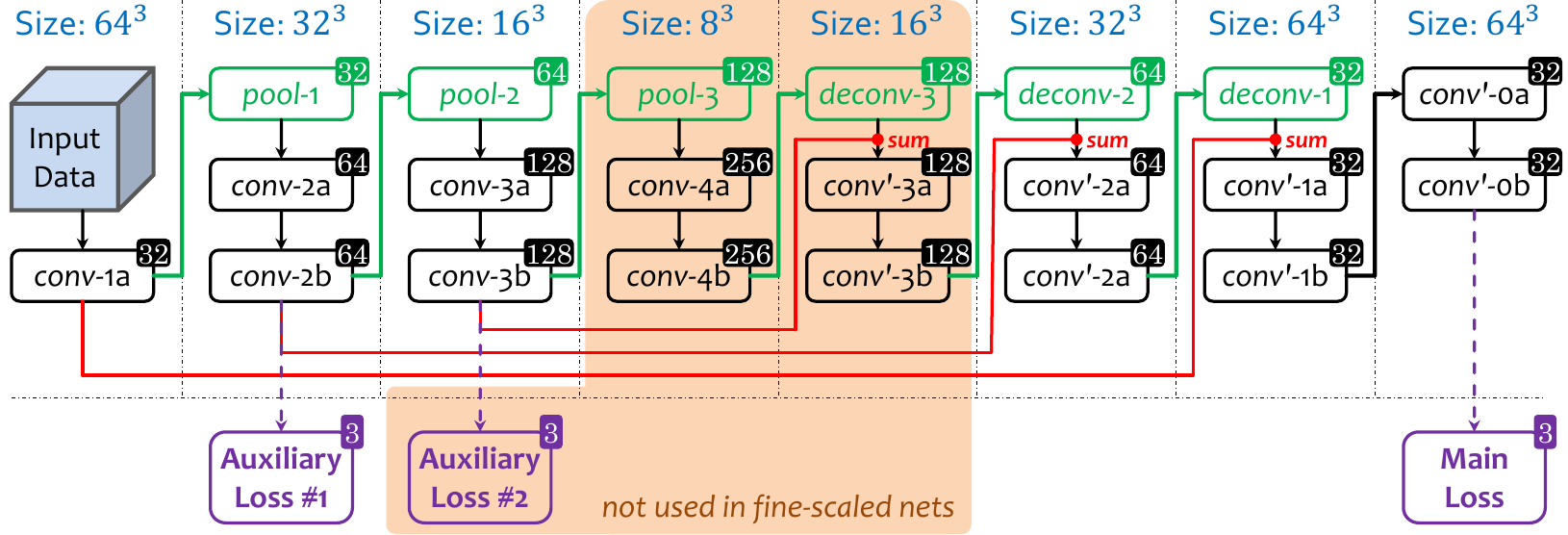}
% \vspace{-0.5cm}
\caption{
    The architecture of our segmentation backbone (best viewed in color). Each rectangle is a layer, green arrows indicate operations changing spatial resolution, and red arrows mean residual connections. We illustrate the situation when the input volume size is $64^3$. If it is $32^3$ or $16^3$, all volumes are shrunk accordingly (to $1/2^3$ or $1/4^3$ of the displayed size). The number at the upper-right corner of each cube is the number of channels. Each convolution uses $3$$\times$$3$$\times$$3$ kernels with $1$ as stride, each pooling $2$$\times$$2$$\times$$2$ with $2$ as stride, and each deconvolution $4$$\times$$4$$\times$$4$ with $2$ as stride. The weight ratio for auxiliary losses \#1, \#2 and main loss is $1:2:5$ for the $64^3$ network, and $1:3$ for the auxiliary loss \#1 and the main loss for the $32^3$ and $16^3$ networks.}
    
    % (down-sampling by $2$) (up-sampling by $2$)
    
    % Batch-normalization and ReLU activation are used after each convolution and deconvolution.
    
    %The loss function works by first up-sampling the cube to the output size via deconvolution, followed by a $1$$\times$$1$$\times$$1$ convolution, and then computing the voxel-wise cross-entropy loss. 

\end{center}
\label{Fig:Structure}
% \vspace{-0.8cm}
\end{figure}

We start with describing the segmentation stage. The tumor region in a pancreas, as shown in Fig.~\ref{Fig:Problem}, can vary in scale, appearance and geometric properties. In particular, the largest tumor in our dataset occupies over one million voxels, but the smallest one has only thousands. This motivates us to train multi-scale networks to deal with such a large variation in scale.

In practice, we train three networks, taking input volumes of $64^3$, $32^3$ and $16^3$ voxels, respectively. Each segmentation network follows an encoder-decoder flowchart shown in Fig.~\ref{Fig:Structure}. It has a series of convolutional layers to learn 3D patterns from training data. Down-sampling and up-sampling are implemented by max pooling and deconvolutional layers, respectively. Following~\cite{zhu2018a}, we introduce deep supervision in the training process, which is implemented by adding several auxiliary losses to intermediate layers, which delivers better performance for the normal and cystic pancreas segmentation in~\cite{zhu2018a}. Deep supervision is considered as a way of incorporating multi-stage visual cues, which constrains intermediate layers and improves the stability of training deep networks. Multi-scale segmentation is complementary to deep supervision, which aims at capturing visual patterns of various scales. As can be seen in experiments, multi-scale segmentation can take advantage of different scales, {\em i.e.}, a large network produces a high specificity, and a small network gives a high sensitivity.

% ~\cite{xie2015holistically}

The training process starts with sampling patches of a specified size. Since the pancreas and the tumor only occupy a small fraction of the entire volume, a random sampling strategy may lead to that only few patches contain pancreas or tumor voxels, and thus the segmentation models are biased towards the background class. To deal with the issue, we sample lots of foreground patches for training the $32^3$ and $16^3$ networks. We first compute the region-of-interest (ROI) by padding a $32$-voxel margin around the minimal 3D bounding box covering the entire pancreas. Within it, we categorize the randomly sampled patches into three types ({\em i.e.}, {\em background}, {\em tumor} and {\em pancreas}) according to the fraction of pancreas and tumor voxels, and make the numbers of training patches of these types o be approximately the same. Data augmentation is performed by randomly flipping patches and rotating by $90^\circ$, $180^\circ$ and $270^\circ$ over three axes.

% ($24$ possibilities in total)

% \footnote{A patch is considered {\em background} if it contains more than $80\%$ background voxels; {\em tumor} if it contains more than $10\%$ tumor voxels; otherwise it is {\em pancreas}.}

We use the same configuration for training these networks. The base learning rate is $0.01$ and decayed polynomially (the power is $0.9$) in a total of $80\rm{,}000$ iterations (the mini-batch size is $16$, $32$ and $128$ for $64^3$, $32^3$ and $16^3$, respectively). The weight decay and momentum are set to be $0.0005$ and $0.9$, respectively.

% \vspace{-0.2cm}
\subsection{Testing: Coarse-to-Fine Segmentation with Post-Processing}
\label{Approach:Testing}

The first goal is to perform the pancreas and tumor segmentation. We first slide a $64^3$ window in the entire CT volume. The spatial stride is $20$ along three axes, which is chosen to have the average testing time for each case within $11$ minutes on a TITAN Xp GPU. Based on the {\em coarse} segmentation, we compute the ROI, {\em i.e.}, the smallest box covering all pancreas and tumor voxels padded by $32$, and crop the CT image accordingly. Then, we scan the ROI with sliding windows of $32^3$ and $16^3$ voxels, and the strides are set to be $10$ and $5$, respectively. We do not run the two small networks on the entire volume because it can easily hallucinate tumors in the background regions. In addition, shrinking the scanning region for the $32^3$ and $16^3$ networks leads to a significant speedup in the testing process. The predictions of three networks are averaged into final segmentation.

Then, based on the segmentation mask, we classify each volume as normal or abnormal. Advised by the radiologists who desire the classification result to be explainable, we do not formulate the classifier $c\!\left(\cdot\right)$ as another deep network, but use a simple, non-parametrized approach to filter out the outliers. We construct a graph on all voxels predicted as {\em normal pancreas} or {\em tumor}. Each voxel is a node, and there exists an edge between the adjacent voxels (each voxel is adjacent to $6$ neighbors). We compute all connected component in the graph. A component is preserved if it is larger than $20\%$ of the maximal connected component, otherwise it is removed, {\em i.e.}, all voxels within this component are predicted as {\em background}. To obtain our final goal, a volume is predicted as PDAC if at least $K$ voxels are predicted as tumor. In practice, we empirically set ${K}={50}$. 

% It is worth noting that our method is not sensitive to $K$. 

% Even setting ${K}={1500}$ only makes subtle changes to our results and observations.

% \vspace{-0.4cm}
\section{Experiments}
\label{Experiments}
% \vspace{-0.4cm}
\subsection{Dataset and Settings}
\label{Experiments:DatasetSettings}
% We also add $103$ normal cases to the training set to suppress false alarms.

We collected a dataset with $303$ normal cases from potential renal donors, as well as $136$ biopsy-proven PDAC cases. Four experts in abdominal anatomy annotated the pancreas and tumor voxels on these data using the Varian Velocity software, and each case was checked by an experienced board-certified Abdominal Radiologist. For a radiologist, an average normal case took $20$ minutes, and an average abnormal case $40$ minutes to segment. Since the abnormal cases are much harder to obtain and annotate than the normal cases, we adopt a $4$-fold cross-validation on our $136$ PDAC scans to have testing results on every abnormal case while we use a hard split of training and testing on our $303$ normal cases. All in all, each training set contains $103$ normal and $102$ abnormal cases where the normal-to-abnormal ratio is close to $1$, and each testing set contains $34$ abnormal and $200$ normal cases. The average size of CT scans is $512$$\times$$512$$\times$$667$.

One goal is to measure the segmentation accuracy by the Dice-S{\o}rensen Coefficient (DSC) between the predicted and the ground-truth tumor sets $\mathcal{Y}$ and $\mathcal{Y}^\star$, {\em i.e.}, ${\mathrm{DSC}\!\left(\mathcal{Y},\mathcal{Y}^\star\right)}={{2\times\left|\mathcal{Y}\cap\mathcal{Y}^\star\right|}/({\left|\mathcal{Y}\right|+\left|\mathcal{Y}^\star\right|})}$. Our main goal is the tumor classification, which involves a tradeoff between sensitivity and specificity. 

% (the fraction of correctly classified abnormal cases) (the fraction of correctly classified normal cases).

% \vspace{-0.2cm}
\subsection{Segmentation Results}
\label{Experiments:Segmentation}

\begin{table}[!btp]
\begin{center}
% \begin{tabular}{|l|R{\colwidthA}||R{\colwidthA}|R{\colwidthA}|R{\colwidthB}||R{\colwidthC}|R{\colwidthC}|}
\begin{tabular}{lccccccc}\toprule
Scale  &                N. Pancreas &                A. Pancreas &                      Tumor &           Misses &             Sensitivity &             Specificity \\
\hline
$64^3$ & $\mathbf{86.9}\pm8.6\%$ & $\mathbf{81.0}\pm10.8\%$ & $\mathbf{57.3}\pm28.1\%$ &         $10/136$ &          $92.7\%$ & $\mathbf{99.0}\%$ \\
\hline
$32^3$ &          $82.0\pm12.2\%$ &          $75.7\pm14.9\%$ &          $53.8\pm26.1\%$ &         $ 7/136$ &          $94.9\%$ &          $96.0\%$ \\
\hline
$16^3$ &          $61.5\pm20.6\%$ &          $64.1\pm20.2\%$ &          $42.5\pm25.6\%$ & $\mathbf{4}/136$ & $\mathbf{97.1}\%$ &          $86.5\%$ \\
\hline
Multi  & $84.5\pm11.1\%$ &          $78.6\pm13.3\%$ &          $56.5\pm27.2\%$ &         $ 8/136$ &          $94.1\%$ &          $98.5\%$ \\
 \bottomrule
\end{tabular}
\end{center}
\caption{
    Comparison of segmentation and classification results by networks of different scales and their combination. From left to right: normal/abnormal pancreas and tumor segmentation accuracy (DSC, $\%$), the number of missing tumors ({\em i.e.}, DSC is $0\%$), and the sensitivity ($=1-\mathrm{\ miss\ rate}$) and specificity.}
    
    % Results of the $32^3$ and $16^3$ networks are based on the predicted bounding box provided by the $64^3$ network.
    % , otherwise the segmentation accuracy is much lower due to a large number of false positives.

\label{Tab:Segmentation}
% \vspace{-0.5cm}
\end{table}

We first summarize the segmentation results in Table~\ref{Tab:Segmentation}, which makes the normal v.s. abnormal classification to be interpretable by segmentation cues. The $64^3$ network achieves reasonable pancreas and tumor segmentation accuracies. The segmentation result of normal pancreas is as high as $86.9\%$, which means that the normal pancreases are easier to segment, as there are often unpredicted changes in shape and geometry in the abnormal cases. As a side comment, the lowest DSC of an abnormal pancreas is $38.4\%$, lower than the number ($44.0\%$) of a normal pancreas. In tumor segmentation, we observe a lower accuracy and a higher standard deviation ($57.3\pm28.1\%$). Except for the $10$ missing cases, we find $20$ more cases with a tumor DSC lower than $30\%$. All these evidences imply the challenging of finding tumors considering their various size, shape and locations. Note that a recent work on the pancreatic cyst segmentation achieves a DSC of $63.4\pm{27.7}\%$~\cite{zhou2017deep}, which is not as hard as the tumor segmentation.

\begin{figure}[!t]
\begin{center}
    \includegraphics[width=12.0cm]{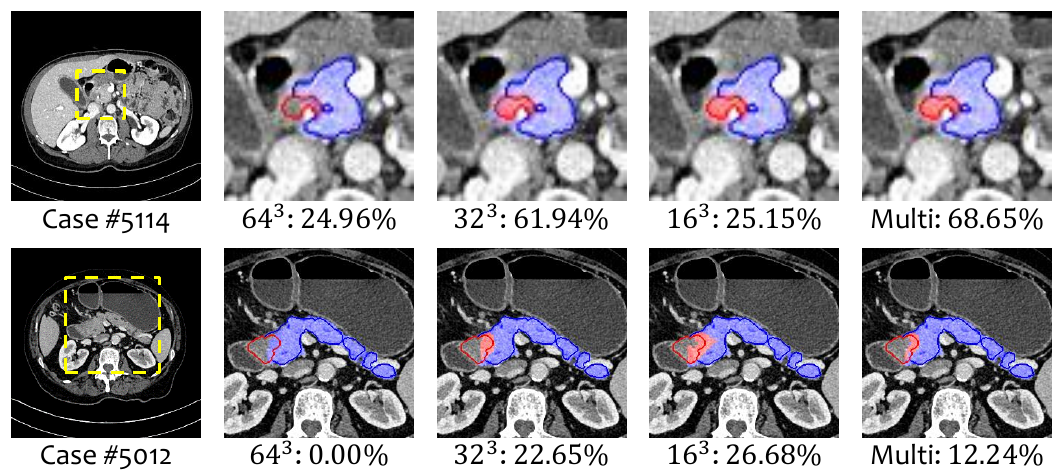}
\end{center}
% \vspace{-0.5cm}
\caption{
    Multi-scale segmentation examples (best viewed in color). Top: a case that all three scales work well, and multi-scale combines them to achieve a higher DSC. Bottom: a failure case in the $64^3$ network, but found by the $32^3$ and $16^3$ networks. The yellow frames indicate the zoomed-in regions, the blue and red contours mark the annotated pancreas and tumor respectively, and the masked regions mark segmentation results.
}
\label{Fig:Examples}
\end{figure}

Going to smaller scales, fewer tumors are missed, though segmentation accuracies become lower. This is the tradeoff between sensitivity and specificity: a network with a smaller input region has the ability to detect tiny regions, but without seeing contexts, it can be easily confused by false positives. Thus, combining multi-scale predictions achieves a balance between sensitivity and specificity. Fig.~\ref{Fig:Examples} shows two examples that benefit from multi-scale segmentation.

We replace our backbone with 3D UNet~\cite{cciccek20163d} and VNet~\cite{milletari2016v} at the $64^3$ scale setting and report their results in Table~\ref{Tab:SegmentationAblation} for comparison. We can find that the three backbones perform roughly similar in terms of the segmentation results. However, our backbone achieves the best results for the sensitivity and specificity.

\begin{table}[!t]
\begin{center}
\begin{tabular}{lccccccc}\toprule
Scale  &                N. Pancreas &                A. Pancreas &                      Tumor &           Misses &             Sensitivity &             Specificity \\
\hline
Ours & $86.9\pm8.6\%$ & $81.0\pm10.8\%$ & $57.3\pm28.1\%$ &         $\mathbf{10/136}$ &          $\mathbf{92.7\%}$ & $\mathbf{99.0\%}$ \\
\hline
UNet &          $\mathbf{87.0\pm8.4\%}$ &          $\mathbf{81.6\pm10.2\%}$ &          $57.6\pm27.8\%$ &         $ 11/136$ &          $91.9\%$ &          $99.0\%$ \\
\hline
VNet &          $86.7\pm8.8\%$ &          $80.6\pm11.4\%$ &          $\mathbf{58.7\pm28.0\%}$ & $10/136$ & $92.7\%$ &          $98.0\%$ \\
% VNet false positives, combine 0267, 0270, 0533, 0635 (0140, 0384, 0652)
% 0140, 0267, 0270, 0533 -- CV1
% 0267, 0270, 0384, 0533, 0635 -- CV2
% 0533, 0635 -- CV3
% 0267, 0270, 0533, 0652 -- CV3
 \bottomrule
\end{tabular}
\end{center}
\caption{
    Comparison of different networks as backbone at the $64^3$ setting.
}
\label{Tab:SegmentationAblation}
% \vspace{-0.8cm}
\end{table}

% \vspace{-0.2cm}
\subsection{Classification Results}
\label{Experiments:Analysis}
% \vspace{-0.1cm}

\begin{figure}[!t]
\begin{center}
    \includegraphics[width=8.4cm]{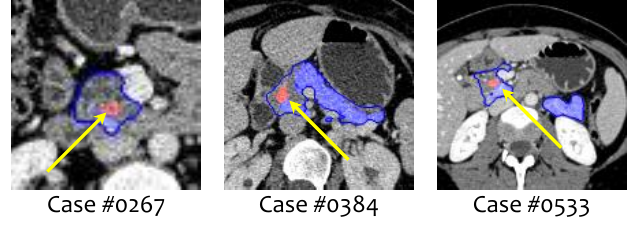}
    \includegraphics[width=3.6cm]{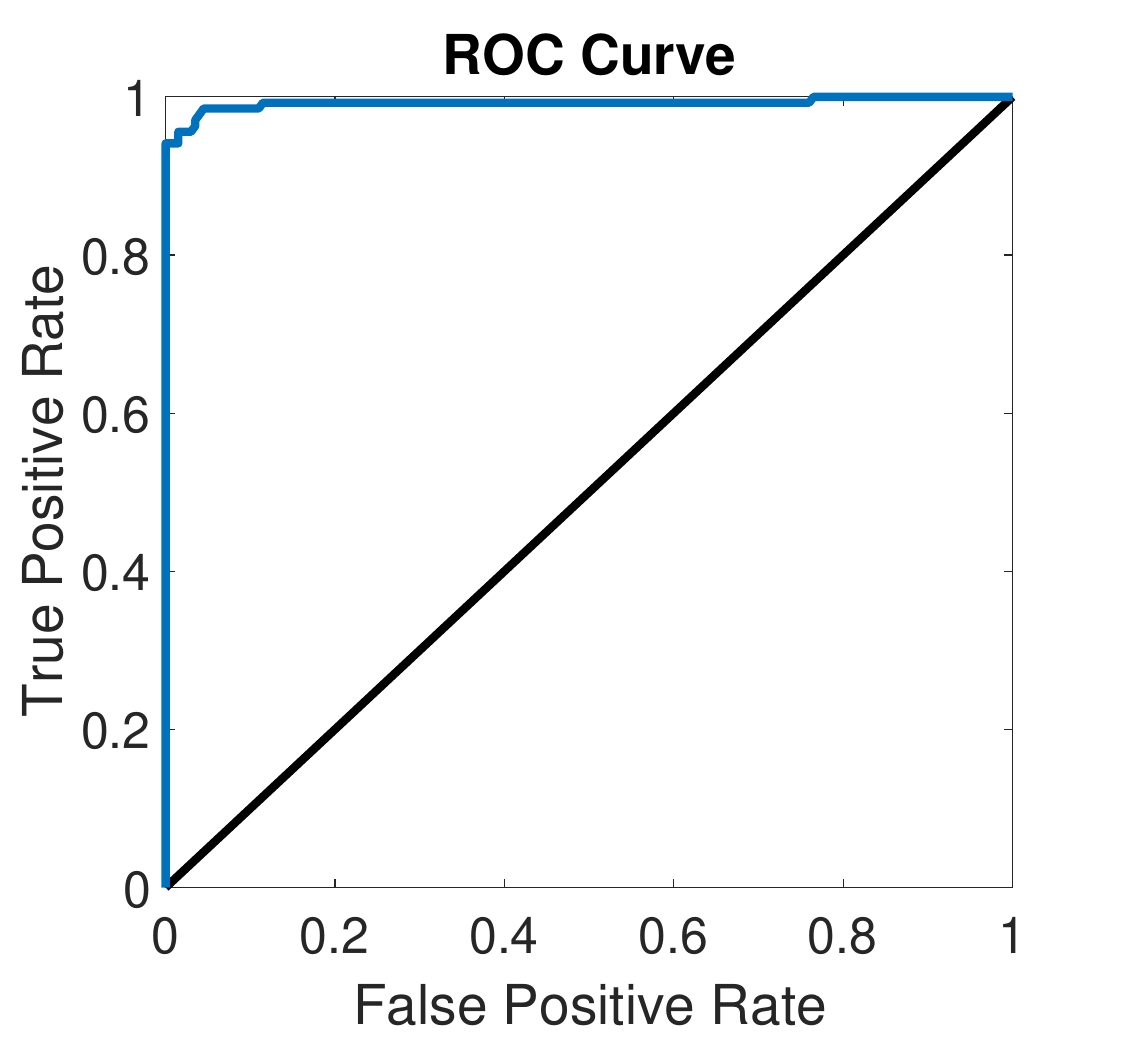}
\end{center}
% \vspace{-0.6cm}
\caption{
    Left: three false alarm examples, in which the blue contour marks the annotated pancreas, and the blue and red regions mark the predicted pancreas and tumor, respectively. We use yellow arrows to indicate the detected tiny ``tumors''. Right: the ROC curve of multi-scale classification. This figure is best viewed in color.
}
\label{Fig:Classification}
% \vspace{-0.7cm}
\end{figure}

Finally, we summarize classification results in Table~\ref{Tab:Segmentation}, which is the crucial goal of making earlier diagnosis possible for doctors. Radiologists care more about a high sensitivity since they don't want to miss a patient who has an abnormal pancreas, which inspires us to adopt a multi-scale strategy to improve the sensitivity while keeping a reasonable specificity. The model with multi-scale information achieves the best overall performance, {\em i.e.}, a sensitivity of $94.1\%$ at a specificity of $98.5\%$. These high scores imply that tumor segmentation provide strong cues for PDAC screening. We show all three false alarms in Fig.~\ref{Fig:Classification}. The radiologists of our team confirmed that $2$ out of these $3$ false positives have focal fatty infiltration in the pancreas corresponding to the detected ``tumors''. Focal fatty infiltration is difficult for radiologists to distinguish from tumor in current clinical practice. In this case, the predicted ``false alarm'' was not normal in view of our radiologists.

By augmenting our segmentation for classification framework with cues from number of predicted tumor voxels since the more voxels predicted as PDAC the more likely this case is abnormal, we can output a confident score for each case, indicating the possibility that this case suffers PDAC. More specifically, a confidence score is computed by a weighted sum of the volume size and the segmentation probability of predicted tumor voxels. By sorting all testing cases according to their confident scores, we obtain a ROC curve of sensitivity and specificity. From the ROC curve, we can make different emphasis to change the tradeoff between sensitivity and specificity, {\em e.g.}, we can achieve a sensitivity of $98.5\%$ at a specificity of $95.6\%$, or a specificity of $99.5\%$ at a sensitivity of $94.1\%$.

% \vspace{-0.3cm}
\section{Conclusion}\label{Conclusions}
% \vspace{-0.2cm}
In this paper, we study an important and challenging task, {\em i.e.}, detecting pancreases diagnosed with PDAC in abdominal CT scans. This topic is crucial in saving lives from pancreatic cancer yet few studied before, possibly due to the lack of data. We propose a {\bf segmentation-for-classification} framework which trains a segmentation network and performs {\bf interpretable} abnormality classification by simply checking the existence of tumor voxels in each testing volume. There are two key points to improve classification accuracy, known as {\bf multi-scale} network training and {\bf coarse-to-fine} testing. To offer a best trade-off between sensitivity and specificity on our own collected dataset containing $303$ normal and $136$ PDAC cases, we achieve a sensitivity of $94.1\%$ at a specificity of $98.5\%$. The strong numbers show the promising direction to make a significant impact in clinics for early detection of pancreatic cancer, which would save lives.

% Our approach enjoys another benefit that producing interpretable predictions, which reduces the workload of human doctors. In the future, we will combine other cues ({\em e.g.}, the shape of the pancreas) into our framework, and explore a joint way of optimizing segmentation and classification.

{\bf Acknowledgements} This work was supported by the Lustgarten Foundation for Pancreatic Cancer Research.

% \vspace{-0.2cm}
\bibliographystyle{splncs03}
\bibliography{typeinst}
\end{document}